\documentclass[letterpaper, 10 pt, conference]{ieeeconf}
\IEEEoverridecommandlockouts
\usepackage{cite}
\usepackage{amsmath,amssymb,amsfonts}
\usepackage{algorithmic}
\usepackage{graphicx}
\usepackage{textcomp}
\usepackage{xcolor}

\title{\LARGE \bf Metrics and Benchmarks for Remote Shared Controllers \\in Industrial Applications}

\author{Claudio Zito$^{1}$, Maxime Adjigble$^{1}$, Brice D. Denoun$^{2}$, Lorenzo Jamone$^{2}$, Miles Hansard$^{2}$ and Rustam Stolkin$^{1}$
}%

\begin{document}

\maketitle

\begin{abstract}
Remote manipulation is emerging as one of the key robotics tasks needed in extreme environments. Several researchers have investigated how to add AI components into shared controllers to improve their reliability. Nonetheless, the impact of novel research approaches in real-world applications can have a very slow in-take. We propose a set of benchmarks and metrics to evaluate how the AI components of remote shared control algorithms can improve the effectiveness of such frameworks for real industrial applications. We also present an empirical evaluation of a simple intelligent share controller against a manually operated manipulator in a tele-operated grasping scenario.  
\end{abstract}

\section{Introduction}

The exposure of humans to hostile work environments can be reduced by means of shared controlled systems, in which the operator remotely controls a robotic platform to perform a task in such environments, e.g.~\cite{bib:fulbright_1995}. One of the first industries which initially introduced the use of shared control systems was the nuclear industry but, over time, many other industries have adopted these technologies including health care, mining, military, firefighting, undersea, construction, and space~\cite{bib:kent_2017}. 

Telepresence is achieved by sensing through digital sensors information about the task environment, and feeding back this information to the human operator on a remote site~\cite{sheridan_1992}. 
The level of automation of the system typically depends on the area of application. Some requires only supervision from the operator, but in the majority of cases it requires direct manual manipulation via a master device. In fact, the current state-of-the-art controllers in extreme environments still heavily relies on the user-in-the-loop for cognitive reasoning, while the robot device merely reproduces the master's movements. However, direct control is typically non-intuitive for the human operator and yields to an increase of his cognitive burden, which has led many researchers to investigate possible alternatives towards more efficient and intuitive frameworks, e.g.~\cite{marturi_2018,bib:fulbright_1995,bib:kent_2017}. 

\begin{figure}[t]
\centering
\includegraphics[width=0.7\columnwidth]{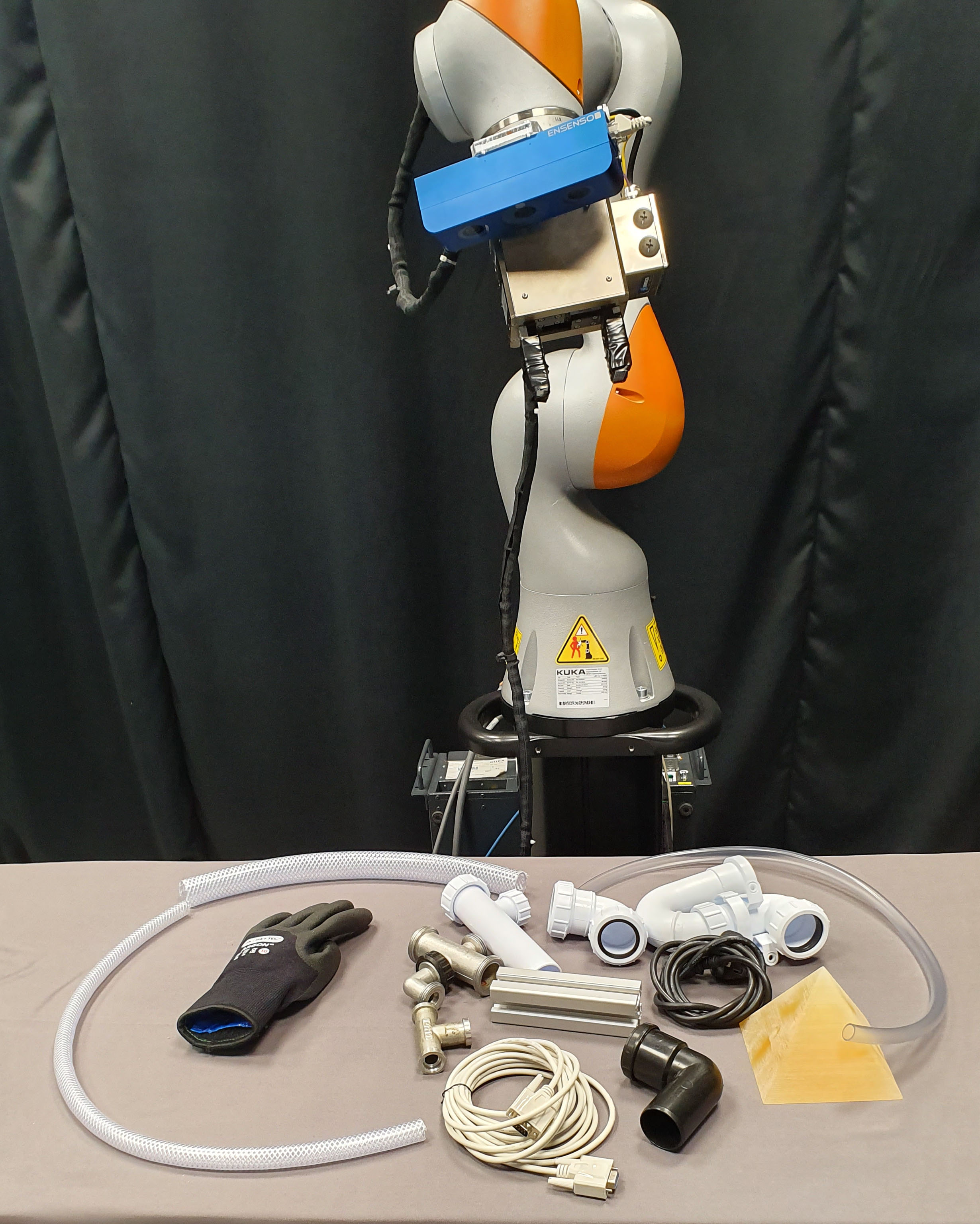}
\caption{The slave robot and an example of industrial objects from our dataset. The robot is an KUKA IIWA equipped with an eye-on-hand Ensenso RGB-D camera and a Schunk Parallel Gripper. The object are randomly selected from different categories: deformable, transparent, metallic, combined, and primitive 3D printed shapes.}
\label{fig:dataset}
\end{figure}

The key insight is to add an active AI component which is context- and user-aware so to make better decision on how to assist the operator. Context-awareness is typically provided by reconstructing and understanding the scene, in terms of the objects the robot has to manipulate. User-awareness is obtained by providing as input the operator's task to the AI component, so to enable a more efficient interpretation of his/her inputs through the master device.   

Despite the advancements in technologies and algorithms in autonomous robotics, e.g.~\cite{kopicki_2015,zito_2019,stuber_2019,zito_2013,zito_w2012,zito_w2013,rosales_2018}, and shared controllers, e.g.~\cite{marturi_2018,bib:fulbright_1995,bib:kent_2017},
many industries has not yet embraced these new approaches, but rather prefer to maintain out-of-date but reliable systems. This is due to a very simple fact: the risks for the operators and the money to be invested are not worth the benefits that a novel approach may have on paper but which has never been properly tested on an uniform and standardised benchmark. Therefore we argue that providing such a benchmark, a benchmark approved and standardised by a consortium of research institutions and industries, will encourage industrial partners to invest of the new technologies, which will lead to a safer and efficient environments.  

The benchmarks and metrics we propose in this paper are designed to evaluate mainly two aspects of a share control algorithm: i) The ability of extrapolating contextual information from sensing the environment to be of any support of the operator, and ii) how the (visual, haptic) feedback are used to influence the operator's response. The combination of these two components should, in fact, improve the task efficiency (number of successful executions of the task), reduce the task effort (how long it will take to execute the task), and robot demand attention (the time the user utilises to interact with the interface instead of focusing on the task at hand).

\begin{figure}[t]
\centerline{\includegraphics[width=0.98\columnwidth]{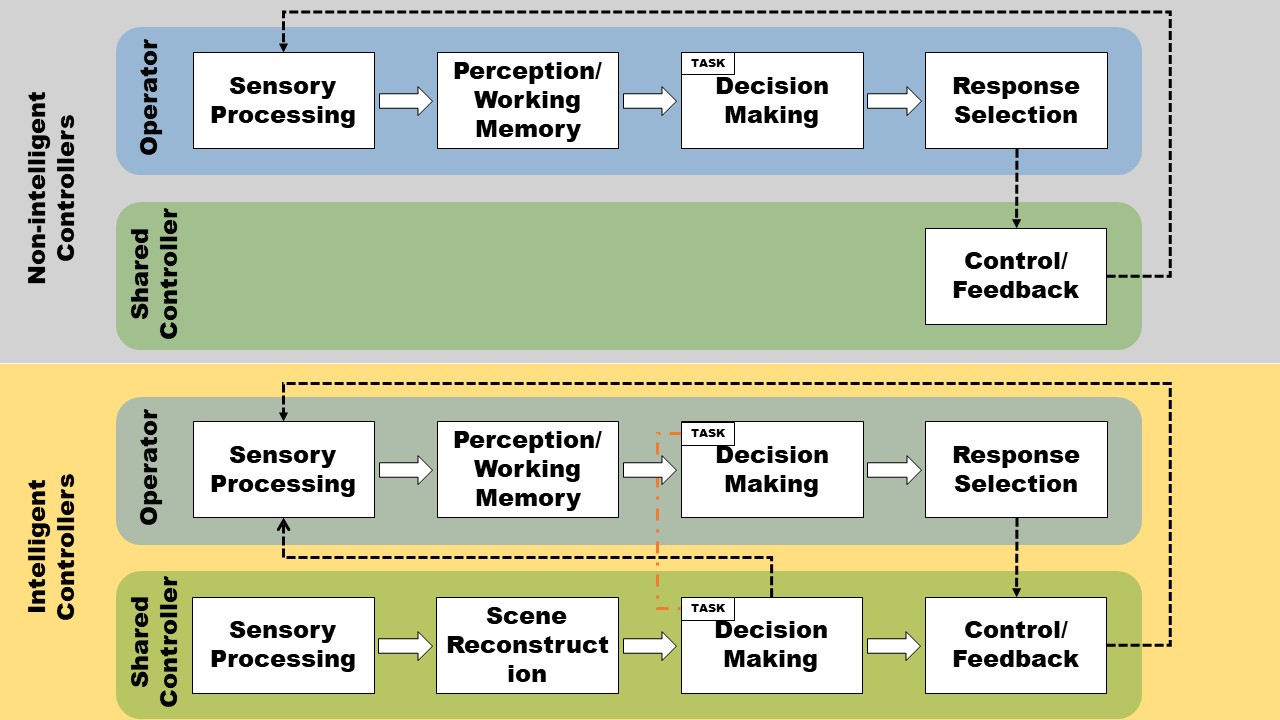}}
\caption{A four-stage model of human/robot information processing. The top row (grey) shows a classical, non-intelligent shared controller, which simply maps the master's movements into the slave's movements. No context- or user-awareness is provided. The bottom row (yellow) shows the same model for an intelligent shared controller that is context-aware thanks to a scene reconstruction of the task space, and user-aware thanks to a shared task which will allowed the algorithm of interpreting the operator's intentions, through his/her responses, in a more helpful way. Modified from~\cite{parasuraman_2000}.}
\label{fig:4stage}
\end{figure}


\section{Background}

 The main task of a shared control interface is to sense appropriate information about the environment and provide them to the human operator. The operator will take his or her decisions upon the information received. Figure~\ref{fig:4stage} (blue model) shows a simple four-stage model of information processing for the operator. The first stage refers to the acquisition and registration of multiple sources of information. The second stage involves conscious perception and retrieving processed information from memory. The third stage is where the decision on how to act is made. The decision is obviously influenced by the task at hand. The fourth and final stage is the implementation of the chosen response, typically as a movement on a master device.
 
 A non-intelligent shared control system would receive the response of the operator via the master device and mimic (the direction of) the movement on the slave robot, as shown in Fig.~\ref{fig:4stage} (top grey region). The shared controller has no knowledge of the working environment and it is not aware of the task at hand. Hence it is not capable to interpret the operator's intentions, and the best it can do is to reproduce the master's movements and provide some low-level, haptic feedback, e.g. vibrations when hitting an external surface. 
 To reproduce the master's movements, the simplest interface would require the operator to control one joint of the manipulator at the time. First the operator manually chooses the appropriate joint to move and then increases or decreases the joint angle via the master device. However, this type of interfaces increase the operator's cognitive burden and are not very efficient especially for complex robots with high degrees of freedom (DOF)~\cite{sheridan_1992}. Other interfaces control the manipulator in Cartesian space, where the master's movements are replicated by the robot's end-effector w.r.t. a chosen world's coordinate frame. Although it is more efficient than controlling single joints, this process still heavily relies on a visual servoing from the operator.  
 
 \begin{figure}[t]
\centerline{\includegraphics[width=0.99\columnwidth]{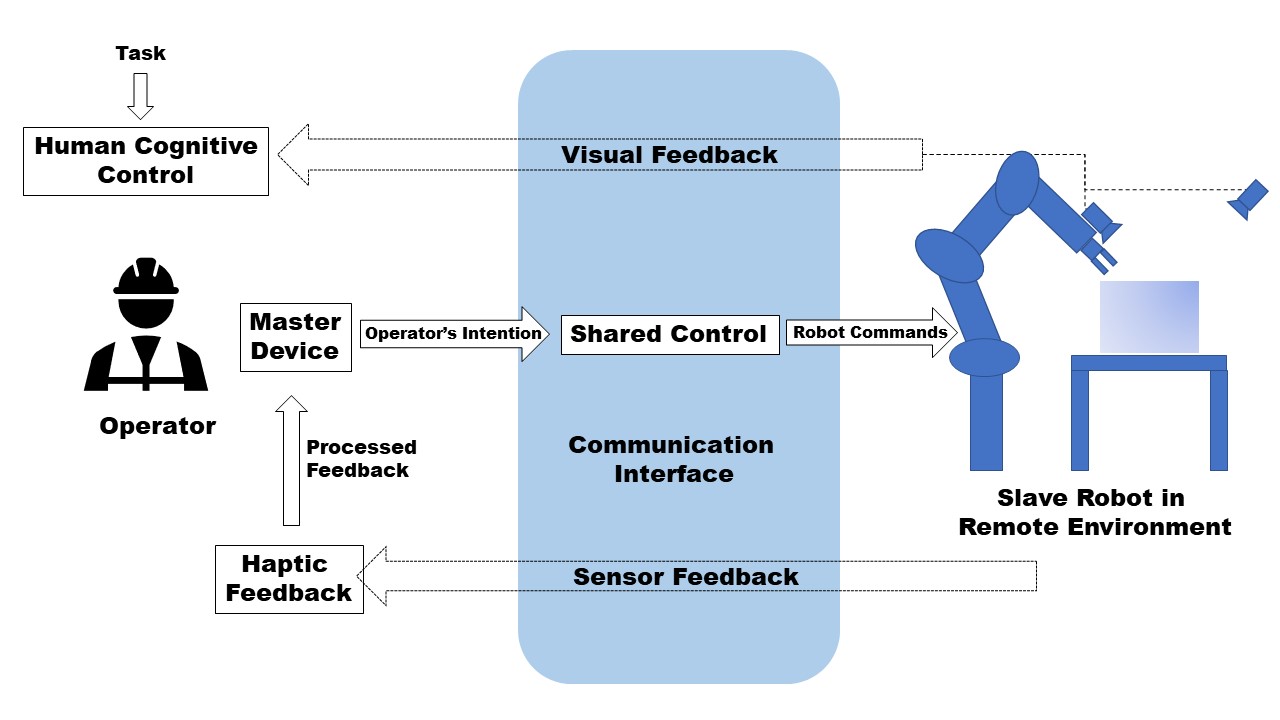}}
\caption{Graphical representation of a remote shared control setup. The operator receives a visual feedback of the task environment (shaded blue area above the tabletop) and controls the slave robot via a master device. The operator input is translated in movements for the slave robot according to the shared control algorithm. Haptic feedback can also be provided to guide the operator according to the shared control algorithm.}
\label{fig:diagram}
\end{figure}

\begin{figure*}[t]
\centering
\includegraphics[width=0.50\columnwidth]{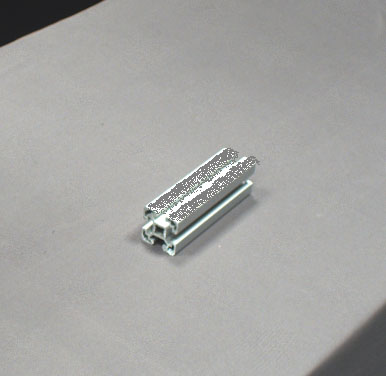}
\includegraphics[width=0.49\columnwidth]{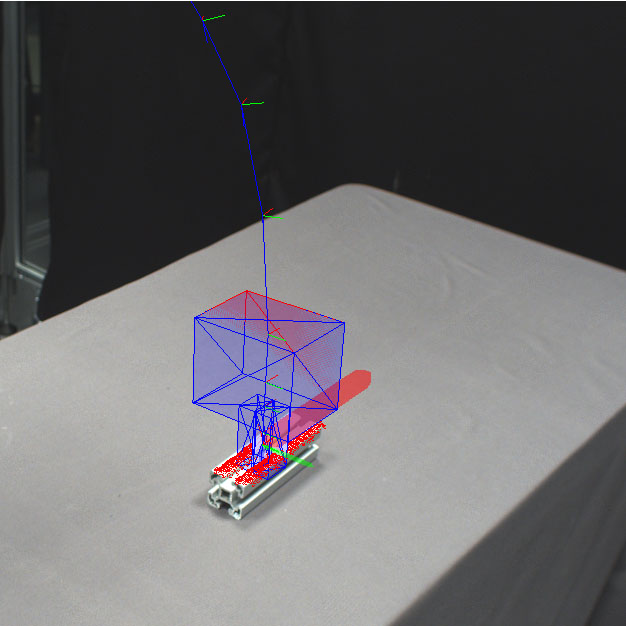}
\includegraphics[width=0.49\columnwidth]{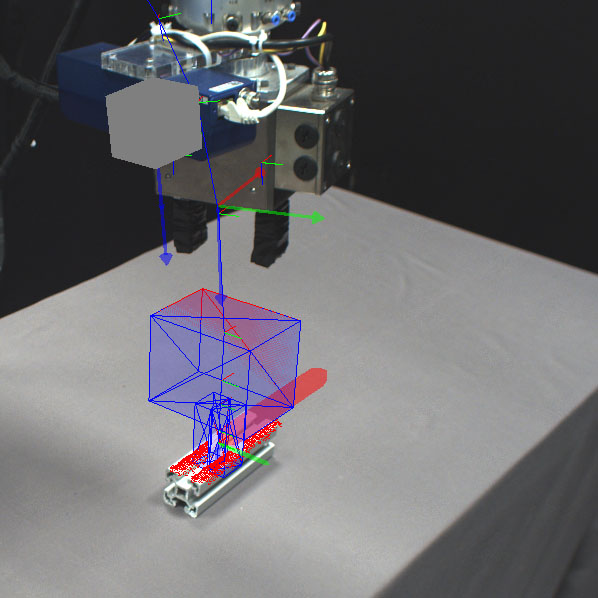}
\includegraphics[width=0.49\columnwidth]{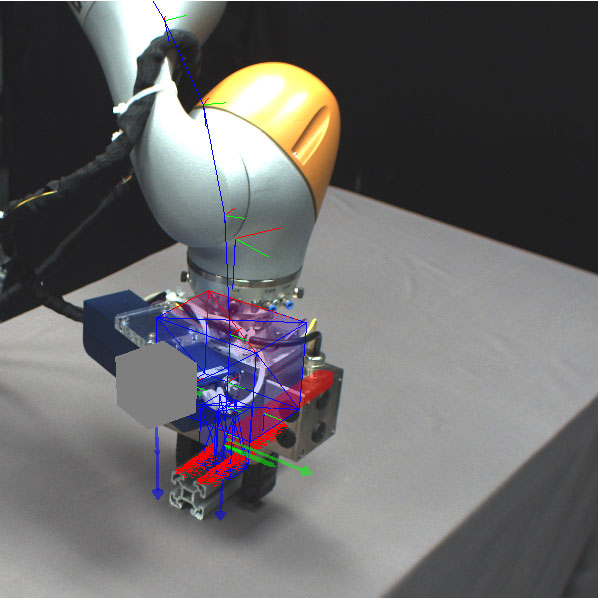}
\caption{An example of an intelligent shared controller. First image on the left: a point cloud is collected to process the workspace. Second image: a grasp suggestion with a relative trajectory is visualised to the operator. Third and fourth images: the operator drives the robot towards the object following the suggested trajectory until the operator decides to send the command to close the gripper and grasp the object.}
\label{fig:reach}
\end{figure*}

 Figure~\ref{fig:4stage} (bottom row, green model) shows the relative four-stage model for an intelligent shared controller and how the operator/robot share such information (black dotted arrows). The first stage involves capturing one or multiple view of the task space. Typically this is done by acquiring RGB-D images from pre-selected poses from the eye-on-hand camera. If multiple images are taken, the system registers them to create a single dense point cloud. The second stage pre-process the point cloud to remove unnecessary surfaces, such as the tabletop, or outliers. A decision is then made given the contextual information available. Assuming that the AI system and the operator share the same task, such as pick and place the objects from the tabletop, the former can assist the decision-making process of the latter by offering available actions. In the pick-and-place example, the system could compute a set of candidate grasping trajectories and visualise them on the augmented feed of the external camera, as shown in Fig~\ref{fig:4stage} by the black dotted arrow from the shared controller's Decision Making module to the operator's Sensory Processing. This would provide a way to influence the operator into select the preferred grasp. Once this process is complete, the operator and the shared controller are working to reach and execute a selected action, of which they are both aware. The system will interpret the operator's commands to assist him/her to accomplish the goal, e.g. moving along the trajectory. Finally, the fourth stage of the shared control involves visual or haptic feedback that can be used to communicate with the human to affect his/her response, e.g. creating a field force on the master haptic device if the robot is moving away from an optimal alignment with the object, jeopardising the grasping success. An example of an intelligent shared control is presented in Fig.~\ref{fig:reach}.
 
\section{Benchmark I: Grasp Efficiency}

\subsection{Objective}
Our main objective in this benchmark is to evaluate how AI algorithms can improve the efficiency of reach and pick up objects in a tele-operated framework. 

\subsection{Dataset}\label{sec:bm1_db}

The dataset is composed of three sets of 3D printed primitive shaped objects, such as cubes, pyramids, and spheres. The first set is designed to offer the best possible scenario, in which the object should be clearly visible from a depth camera and made of non-slippery material so not to challenge the grasps.
The second set of primitive shapes should be made of transparent plastic to challenge the perception abilities of the system, and therefore the ability of the algorithm to provide a candidate grasp. The third set should be covered with a shiny and low-friction material (e.g. wax) so to challenge the perception as well as the grasping execution.

\subsection{Task}

We aim to evaluate if the algorithm can lead the operator to robustly grasps objects. Each of the object will be presented to the robot attached to a base via a spring. By lifting the object of ten centimeters the external force generated by the spring will challenge the grasp.

\subsection{Metrics}\label{sec:bm1_mx}

Three metrics will be measured for this benchmark, as follows.

\begin{enumerate}
    \item Task efficiency: the efficiency of the human-robot team to perform the task, measured as pick-and-hold success rate.
    \item Task effort: the required time to complete the task.
    \item Robot demand attention: the total amount of time that the operator has to spend to align the end-effector with the object before making a grasp.  
\end{enumerate}

\section{Benchmark II: Pick \& Place}

\subsection{Objective}
Our main objective for this benchmark is to evaluate how AI algorithms can assist the operator in pick and place tasks. 

\subsection{Dataset}\label{sec:bm2_db}

The dataset is composed of three sets of objects, as follows.

\begin{itemize}
\item Soft \& deformable: such as objects as gloves, wires, plastic tubes.

\item Metallic-like \& slippery: this are objects that can be 3D printed and sprayed so to result difficult to be sense by a depth camera. Additionally the surface will be spread with a substance to reduce the friction and thus making the grasping more challenging.

\item Composed objects: such as L-joint tubes or a net filled up with bolts. This dataset will challenge the grasping algorithm by adding a dynamic component to the testbed.
\end{itemize}

\subsection{Task I}\label{sec:bm2_tk1}

We will present a single object from the dataset to the robot. A random position in the robot's workspace will be chosen, but maintained constant through the comparisons with the algorithms.
A point cloud from a pre-defined single view will be collected from the eye-in-hand camera.
The operator will need to pick the object and place it to a basket area situated at the side of the workspace.

\subsection{Task II}\label{sec:bm2_tk2}

We will present a clutter scene with objects from the same class of the dataset to the robot. A random position in the robot's workspace will be chosen for each object, but maintained constant through the comparisons with the algorithms.
A point cloud from a pre-defined single view will be collected from the eye-in-hand camera.
The operator will need to pick an object at the time and place it to a basket area situated at the side of the workspace until there are no more objects.

\subsection{Task III}\label{sec:bm2_tk3}

Similarly to Task II described in Sec~\ref{sec:bm2_tk2}, we will present a clutter scene to the robot, but we will allow sampling across categories. Again a random position in the robot's workspace will be chosen for each object, but maintained constant through the comparisons with the algorithms.
A point cloud from a pre-defined single view will be collected from the eye-in-hand camera.
The operator will need to pick an object at the time and place it to a basket area situated at the side of the workspace until there are no more objects.

\subsection{Metrics}\label{sec:bm2_mx}

Three metrics will be measured for this benchmark, as follows.

\begin{enumerate}
    \item Task efficiency: the efficiency of the human-robot team to perform the task, measured as pick-and-place success rate.
    \item Task effort: the required time to complete the task.
    \item Robot demand attention: the total amount of time that the operator has to spend to align the end-effector with the object before making a grasp.  
\end{enumerate}

\section{Benchmark III: Assembly}

\subsection{Objective}

Our main objective for this benchmark is to evaluate how AI algorithms can assist the operator in manipulating objects. 

\subsection{Dataset}\label{sec:bm3_db}

The dataset is composed of a set of 3D printed primitive shapes and a peg board for the respective shapes.

\subsection{Task}

A CAD model of the peg board will be available to the algorithm and its pose will be kept fixed throughout the experiments. We will present a single object from the dataset to the robot in a pre-defined region of the dexterous workspace of the robot manipulator.
A point cloud from a pre-defined single view will be collected from the eye-in-hand camera to localise the object in an expected region.
The operator will need to pick the object and push it through the correct hole in the peg board.

\subsection{Metrics}\label{sec:bm3_mx}

Three metrics will be measured for this benchmark, as follows.

\begin{enumerate}
    \item Task efficiency: the efficiency of the human-robot team to perform the task, measured as peg-in-the-hole success rate.
    \item Task effort: the required time to complete the task.
    \item Robot demand attention: the total amount of time that the operator has to spend to align the grasped object with the correct hole on the peg board.
\end{enumerate}

\section{Experiments}

\subsection{Setup}

To simulate many real-world environments in which the user is back in safety away from where the robot operates, we proposed a setup in which the robot manipulator and its workspace is hidden from the user. The user will be provided with a 2D/3D visual feedback from a set of cameras and a mean to control the manipulator.
As shown in Fig.~\ref{fig:diagram}, two cameras will be placed in the remote environment. One RGB-D camera is mounted on the wrist of the robot for a so called \emph{eye-on-hand} prospective. A second 4K RGB camera will be placed such that the entire dexterous workspace of the robot will be visible. The second camera should be calibrated with the framework so that it is possible to augment the feed from the camera with critical information from the algorithm, e.g. a candidate grasp location or trajectory. 
A 6D haptic device is used as master controller and to receive haptic feedback. 
The robot manipulator is composed by a 7 degrees of freedom (DOF) robot arm mounted on a solid base and equipped with a parallel gripper.


\subsection{Baseline Controller}
Our framework also employs a standard manually tele-operated controller in Cartesian space, as in Fig~\ref{fig:4stage} (grey section), which is commonly used in industrial setups. The master's movements are directly maps into the slave's end-effector movements in 6D with respect to a chosen world's reference frame. The robot gripper is controlled with an extra ON/OFF button for opening and closing. No haptic feedback is provided to the operator. This will provide a beseline of comparison for all the intelligent shared controller algorithm that will employ our proposed benchmarks.

\subsection{Benchmark I}

Each algorithms, included the baseline controller, will be tested in the same conditions. For each object in the dataset described in Sec.~\ref{sec:bm1_db}, ten random positions that span the entire workspace will be selected and recorded. Each algorithm will be tested five times for each pose. This would provide a set of fifty trials per object, and it guarantees that the performance is not biased by the configuration of the object. The metrics presented in Sec.~\ref{sec:bm1_mx} will be computed at the end of the experiments as average per class (i.e. standard 3D printed object, transparent, and shiny-slippery) and across classes.

\subsection{Benchmark II}

Again, each algorithms, included the baseline controller, will be tested in the same conditions. For \emph{Task I} (Sec~\ref{sec:bm2_tk1}) each object in the dataset (Sec.~\ref{sec:bm2_db}) will be placed in ten random positions that span the entire workspace and the positions recorded. For \emph{Task II} (Sec~\ref{sec:bm2_tk2}) and \emph{Task III} (Sec~\ref{sec:bm2_tk3}) ten random clutter scenes with ten objects each will be selected and recorded.
Each algorithm will be tested five times for each object or scene. This would provide a set of fifty trials per object/scene, and it guarantees that the performance is not biased by the configuration of the object or the selected objects in the scene. The metrics presented in Sec.~\ref{sec:bm2_mx} will be computed for each task individually at the end of the experiments as average per class (i.e. soft \& deformable, metallic-like \& slippery, and composed) and across classes.

\subsection{Benchmark III}

Similarly, each algorithms, included the baseline controller, will be tested in the same conditions. Each trial will be composed by twelve objects (i.e. three for each shape) from the dataset described in Sec.~\ref{sec:bm3_db}. The objects will be presented to the operator in a random order which will be recorded. Ten trials per algorithm will be performed. The metrics presented in Sec.~\ref{sec:bm3_mx} will be computed as an average over the ten trials. 

\begin{figure}[t]
\centerline{\includegraphics[width=0.99\columnwidth]{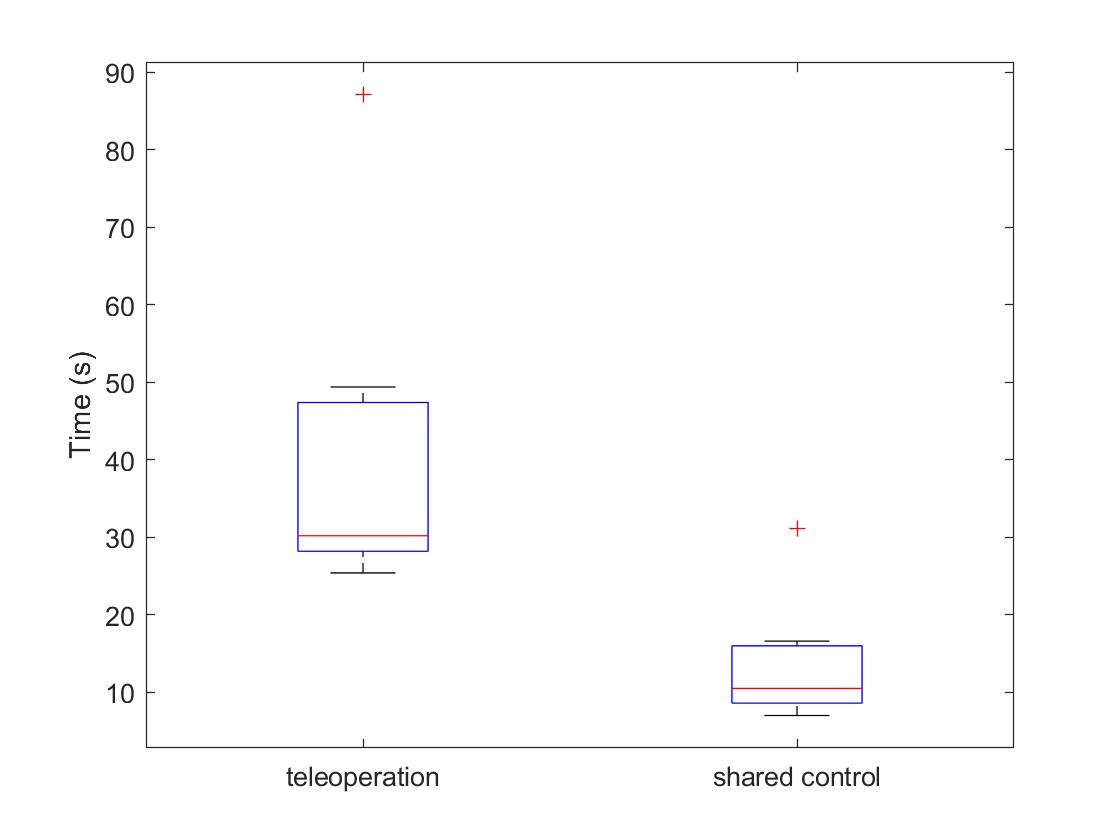}}
\caption{Time effort (in seconds) for teleoperating a robot manipulator for a grasping task. We tested 10 participants. The plot shows the average completion time to reach and grasp 5 objects with standard Cartesian controller (blue) and a simple shared controller (red). The empirical results show that the shared controller outperforms the baseline controller in guiding the robot towards the grasp.}
\label{fig:timechart}
\end{figure}

\section{Conclusion and Future Work}

We have proposed a set of benchmarks and metrics to evaluate algorithms for intelligent remote control in industrial applications. Our proposed design address mainly two aspects of the algorithms: i) how they map the operator's response into a more intuitive controller, and ii) how information are shared/feedback with the operator.
To demonstrate the benefits of these two components, we present an empirical evaluation between a non-intelligent Cartesian controller and a simple algorithm that follows a given grasp trajectory for assisting the operator in pick-up tasks. The intelligent algorithm provides a visual feedback of the proposed grasping trajectory and a haptic feedback to constrain the movements of the operator in the direction of the selected grasp. The operator is only able to move forward or backward along the direction of the selected trajectory, while all the other directions of movement are blocked. The algorithm reduces the operator's cognitive burden in the sense that with a degree of freedom in the master device, the operator is controlling the slave robot in a 6D trajectory. 
Fig. \ref{fig:timechart} shows some preliminary results that show how effective an intelligent shared control can be. We measured for ten participants the task effort in a tele-operated grasping scenario. We asked each participant to guide the slave robot into grasping a single object presented on the tabletop in front of the robot. Before enabling the telo-operation, the tabletop was captured by the eye-on-hand depth camera from a single frontal view, as shown in Fig.~\ref{fig:dataset}. We tested a total of five industrial objects placed in a random position, but constant over the comparison between the algorithms and for each participant. The objects have been randomly selected from the one shown in Fig~\ref{fig:dataset}. The empirical results show that in terms of completion time the shared control outperforms the baseline controller.

As future work, we aim to implement all the benchmarks and metrics in an open source framework and to release the relative datasets of objects. Each dataset will be composed of dense and complete RGB-D scan of each object.  

\bibliographystyle{plain}
\bibliography{RSS2019w}

\begin{thebibliography}{10}

\bibitem{bib:fulbright_1995}
Ron Fulbright and Larry~M. Stephens.
\newblock Swami: An autonomous mobile robot for inspection of nuclear waste
  storage facilities.
\newblock {\em Autonomous Robots}, 2(3):225--235, 1995.

\bibitem{kopicki_2015}
Marek Kopicki, Renaud Detry, Maxime Adjigble, Rustam Stolkin, Ales Leonardis,
  and Jeremy~L. Wyatt.
\newblock One-shot learning and generation of dexterous grasps for novel
  objects.
\newblock {\em The International Journal of Robotics Research}, 35(8):959--976,
  2015.

\bibitem{bib:kent_2017}
Kent~Yee Lui, Hyunjun Cho, ChangSu Ha, and Dongjun Lee.
\newblock First-person view semi-autonomous teleoperation of cooperative
  wheeled mobile robots with visuo-haptic feedback.
\newblock {\em The International Journal of Robotics Research},
  36(5-7):840--860, 2017.

\bibitem{marturi_2018}
Naresh Marturi, Alireza Rastegarpanah, Chie Takahashi, Maxime Adjigble, Rustam
  Stolkin, Sebastian Zurek, Marek Kopicki, Mohammed Talha, Jeffrey~A Kuo, and
  Yasemin Bekiroglu.
\newblock Towards advanced robotic manipulation for nuclear decommissioning: A
  pilot study on tele-operation and autonomy.
\newblock In {\em Proc. of International Conference on Robotics and Automation
  for Humanitarian Applications ({RAHA})}, 2016.

\bibitem{parasuraman_2000}
Raja Parasuraman, Thomas~B. Sheridan, and Christopher~D. Wickens.
\newblock A model for types and levels of human interaction with automation.
\newblock In {\em IEEE Transactions on systems, man, and cybernetics-Part A:
  Systems and Humans}, pages 286--297, 2000.

\bibitem{rosales_2018}
Carlos~J. Rosales, Federico Spinelli, Claudio Zito, Marco Gabiccini, and
  Jeremy~L. Wyatt.
\newblock Gpatlasrrt: A tactile exploration strategy for novel object shape
  modeling.
\newblock {\em International Journal of Humanoid Robotics, Special Issue
  'Tactile perception for manipulation: new progress and challenges'}, 15(1),
  2018.

\bibitem{sheridan_1992}
Thomas~B. Sheridan.
\newblock {\em Telerobotics, Automation, and Human Supervision Control}.
\newblock PhD thesis, Cambridge, MA, 1992.

\bibitem{stuber_2019}
Jochen St{\"u}ber, Claudio Zito, and Rustam Stolkin.
\newblock Let's push things forward: A survey on robot pushing.
\newblock {\em arXiv preprint arXiv:1905.05138 [cs.RO] (cs.AI)}, 2019.

\bibitem{zito_w2013}
Claudio Zito, Marek Kopicki, Rustam Stolkin, Christopher Borst, Florian
  Schmidt, Maximo~A. Roa, and Jeremy~L. Wyatt.
\newblock Sequential re-planning for dextrous grasping under object-pose
  uncertainty.
\newblock In {\em Workshop on Manipulation with Uncertain Models, Robotics:
  Science and Systems ({RSS})}, 2013.

\bibitem{zito_2013}
Claudio Zito, Marek Kopicki, Rustam Stolkin, Christopher Borst, Florian
  Schmidt, Maximo~A. Roa, and Jeremy~L. Wyatt.
\newblock Sequential trajectory re-planning with tactile information gain for
  dextrous grasping under object-pose uncertainty.
\newblock In {\em {IEEE} Proc. Intelligent Robots and Systems ({IROS})}, 2013.

\bibitem{zito_2019}
Claudio Zito, V.~Ortienzi, M.~Adjigble, Marek Kopicki, Rustam Stolkin, and
  Jeremy~L. Wyatt.
\newblock Hypothesis-based belief planning for dexterous grasping.
\newblock {\em arXiv preprint arXiv:1903.05517 [cs.RO] (cs.AI)}, 2019.

\bibitem{zito_w2012}
Claudio Zito, Rustam Stolkin, Marek Kopicki, Massimiliano Di~Luca, and
  Jeremy~L. Wyatt.
\newblock Sequential re-planning for dextrous grasping under object-pose
  uncertainty.
\newblock In {\em Proc. Workshop on Beyond Robot Grasping: Modern Approaches
  for Dynamic Manipulation. Intelligent Robots and Systems ({IROS})}, 2012.

\end{thebibliography}

\end{document}